\documentclass[letterpaper, 10 pt, conference]{ieeeconf}  % Comment this line out if you need a4paper
\usepackage{cite}         
\usepackage[colorlinks=true, citecolor=blue, linkcolor=blue]{hyperref}

\IEEEoverridecommandlockouts                              % This command is only needed if 
\usepackage{amsmath} % assumes amsmath package installed
\usepackage{amssymb}  % assumes amsmath package installed

\usepackage{graphicx}
\usepackage{subcaption}
\usepackage[export]{adjustbox}

\usepackage{tabularx}
\usepackage{multirow}

\usepackage{comment}
\renewcommand{\arraystretch}{1.2}

\usepackage{booktabs}
\usepackage[table]{xcolor}

\usepackage{booktabs}
\usepackage{pifont} % for checkmark and xmark

\title{\LARGE \bf
MyGO-Splat: Multi-Objective Closed-Loop Geometric\\Feedback for RGB-Only Gaussian SLAM
}

\author{Fan Zhu$^{1,2,3}$, Ziyu Chen$^{1,2}$*, Zhenjun Zhao$^{4}$, Zhisong Xu$^{5}$, Hui Zhu$^{1,2}$, \\
	Mingrui Li$^{6}$, Chunmao Jiang$^{1,2}$, Javier Civera$^{4}$	
	\thanks{This work was supported by China Postdoctoral Science Foundation under Grant 2025M781668; Anhui Province Key Research and Development Plan under Grant 202423k09020037.}
	\thanks{$^{*}$ Corresponding author}%
	\thanks{$^{1}$ HFIPS, Chinese Academy of Sciences, Hefei, China.}%
	\thanks{$^{2}$ University of Science and Technology of China, Hefei, China.}%
	\thanks{$^{3}$ Tohoku University, Sendai, Japan}%
	\thanks{$^{4}$ University of Zaragoza, Zaragoza, Spain.}
	\thanks{$^{5}$ University of Tokyo, Tokyo, Japan.}%
	\thanks{$^{6}$ Dalian University of Technology, Dalian, China.}
}

\begin{document}

\maketitle
\thispagestyle{empty}
\pagestyle{empty}

%%%%%%%%%%%%%%%%%%%%%%%%%%%%%%%%%%%%%%%%%%%%%%%%%%%%%%%%%%%%%%%%%%%%%%%%%%%%%%%%
\begin{abstract}
Real-time monocular Simultaneous Localization and Mapping (SLAM) fundamentally suffers from scale ambiguity and a lack of geometric self-correction. While 3D Gaussian Splatting (3DGS) enables high-fidelity rendering, existing RGB-only systems remain open-loop because depth priors are injected into mapping but refined geometry cannot effectively regulate tracking drift. We present MyGO-Splat, a closed-loop Gaussian SLAM framework that analytically rasterizes Gaussian primitives into pixel-wise depth and surface normals, allowing the map to actively supervise camera pose optimization.
To bridge monocular priors and scale consistency, our framework introduces scale-aware adaptive alignment that projects foundation-model depth estimates into the globally optimized Gaussian space, forming a self-correcting cycle for scale feedback. Extensive evaluations show that this closed-loop design improves scale stability and appearance-geometry consistency, achieving performance comparable to RGB-D methods while using only monocular input.
\end{abstract}

%%%%%%%%%%%%%%%%%%%%%%%%%%%%%%%%%%%%%%%%%%%%%%%%%%%%%%%%%%%%%%%%%%%%%%%%%%%%%%%%
\section{Introduction}

Real-time Simultaneous Localization and Mapping (SLAM) aims to jointly estimate the motion of a camera and reconstruct a globally consistent representation of a scene from sensor observations, which is the foundation for driving robots towards embodied and spatial intelligence~\cite{Slamhandbook}. However, for monocular systems, the absence of metric depth measurements introduces an inherent ambiguity regarding scale, which makes long-term geometric consistency fundamentally fragile~\cite{sage,igelio}. Although local bundle adjustment reduces the reprojection error, scale drift and depth distortion cannot be corrected without external constraints~\cite{advances3dcv}. As a result, numerous systems for RGB-only SLAM operate in a loosely coupled or effectively open-loop manner: the tracking thread estimates poses based on short-term correspondences, the mapping thread refines geometry using multi-view supervision, yet this refined geometry rarely participates in the regulation of the tracking process itself. This structural limitations prevents monocular SLAM from achieving reliable self-correction of the geometry.

\begin{figure}[!tbp]
	\centering
	\includegraphics[width=\linewidth]{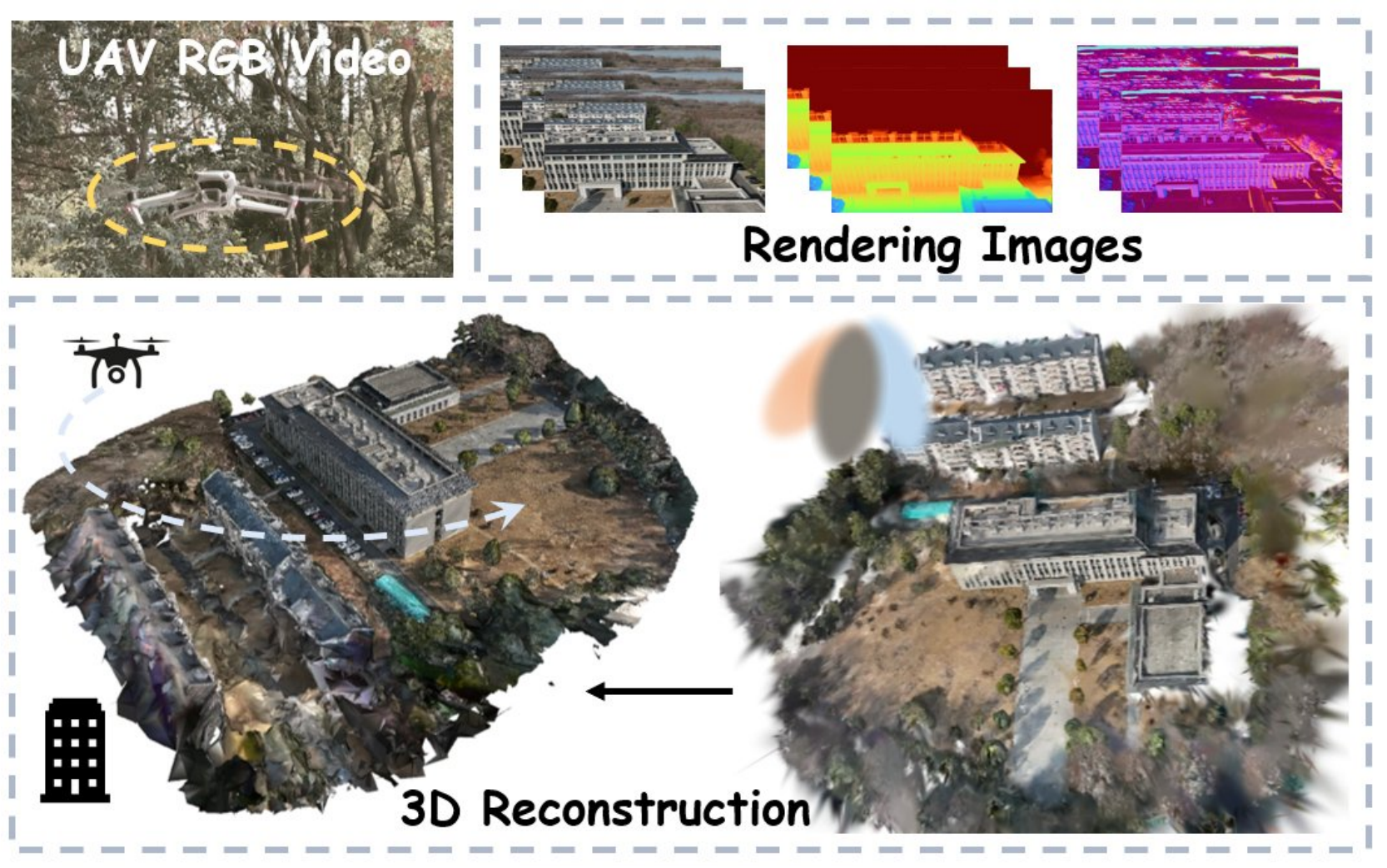}
	\caption{\textbf{MyGO-Splat results.} Our method receives monocular streams and renders high-quality RGB, depth, and normal images, estimating globally consistent representations of the geometry and appearance of a scene.}
	\label{fig:abstract}
	\vspace{-6pt}
\end{figure}

The recent emergence of neural scene representations~\cite{Nerf}, particularly 3D Gaussian Splatting (3DGS)~\cite{3dgs}, has significantly improved the quality of real-time rendering and the expressiveness of maps~\cite{ULF-Loc, PanoImager}. By representing scenes by explicit Gaussian primitives, these methods enable efficient rasterization and high-fidelity reconstruction of their appearance. Several Gaussian-based SLAM systems~\cite{Splatam,MonoGS,Photo-slam,HI-SLAM2} have demonstrated promising results in dense mapping and photorealistic rendering. Nevertheless, most existing approaches rely on accurate depth sensors or treat monocular depth priors as auxiliary supervision during the mapping process. In RGB-only settings, depth predictions from feed-forward models~\cite{Dust3r,vggt} often suffer from scale inconsistency and inter-frame bias. When such priors are directly injected into the optimization of Gaussians, the resulting map may exhibit floaters, distorted surfaces, or inconsistent scale. More critically, the optimized geometry of the Gaussians typically remains passive: although it improves the quality of rendering, it does not actively constrain or correct the frontend for tracking. Consequently, the overall pipeline remains open-loop, and scale drift persists.

To overcome the structural limitation of open-loop Gaussian SLAM, we reformulate the system with a closed-loop geometric feedback framework that upgrades the Gaussian map from a passive renderer to a geometrically active component for tracking supervision. We introduce an analytical rasterization mechanism that derives pixel-wise depth and surface normals from anisotropic Gaussian primitives by modeling ray--Gaussian intersection geometry, enabling multi-view consistent geometric fields rather than center-projection approximations.

In addition, we develop a scale-aware geometric feedback strategy that projects monocular depth priors from 3D vision foundation models into a globally optimized Gaussian coordinate space maintained via loop closure with visual similarity matching. The refined Gaussian-rendered depth is periodically reinjected into tracking optimization, replacing unstable per-frame pseudo-depth with multi-view consistent constraints and progressively suppressing scale drift without external depth sensors.

Our complete system integrates feed-forward tracking, loop closure with global bundle adjustment, closed-loop geometric feedback, and geometric-enhanced multi-objective optimization within a unified closed-loop framework. By jointly enforcing consistency between analytical rasterized geometry, aligned depth priors, and rendered appearance, MyGO-Splat achieves appearance--geometry coherent reconstruction from RGB-only input while maintaining real-time performance.

The contributions of this work are summarized as follows:
\begin{itemize}
	\item We propose a multi-objective closed-loop geometric feedback method for RGB-only Gaussian SLAM, where the Gaussian map actively supervises the tracking geometry through multi-view consistent rendered depth, enabling progressive suppression of scale drift without requiring depth sensors.
	
	\item We introduce an analytical rasterization formulation for anisotropic Gaussian primitives, which derives pixel-wise depth and surface normals directly from the Gaussian representation, transforming it from a rendering-oriented model into a geometrically expressive and optimization-aware scene representation.
	
	\item We develop a scale-aware adaptive alignment mechanism that continuously projects monocular foundation-model depth into a globally optimized Gaussian space, forming a self-correcting scale feedback cycle and enforcing appearance--geometry consistency under RGB-only input.
\end{itemize}
\section{Related Works}
\subsection{Traditional Dense Visual SLAM}
Traditional dense visual SLAM~\cite{Bundlefusion} recovers dense scene structure while estimating camera motion by minimizing photometric or geometric residuals. These methods typically use explicit representations such as point clouds, voxel hashing, and truncated signed distance functions (TSDF) for reconstruction. However, such frameworks rely on photo-consistency and are vulnerable to geometric drift or reconstruction voids under textureless regions, dynamic lighting, or rapid camera motion. Furthermore, their mapping capability in complex indoor environments is constrained by discrete point cloud representations, making it difficult to generate high-fidelity appearance and smooth geometric surfaces.

\subsection{Neural and 3DGS SLAM}
The emergence of neural implicit representations introduces a new paradigm for scene reconstruction. Methods such as iMAP~\cite{Imap} and NICE-SLAM~\cite{Nice-slam} use coordinate networks to model continuous volume density and color fields, demonstrating strong capability in geometric interpolation and completion. More recently, 3DGS has shown promise for real-time dense SLAM due to its explicit Gaussian representation and efficient rasterization. Following the pioneering work of ~\cite{Splatam,Photo-slam,MonoGS}, subsequent research has focused on improving execution speed~\cite{gs-icp,Rtg-slam}, rendering quality~\cite{mmdslam, Segs-slam}, geometric extraction~\cite{FGO-SLAM}, and system robustness~\cite{SLAM-X,GARAD-SLAM,dygs}. Despite these advances, many methods depend on accurate geometric input from depth sensors for tracking and mapping, limiting the broader application of 3DGS-based systems. In contrast, we introduce a rasterized Gaussian geometry approach that jointly optimizes pseudo-depth and rendered depth to achieve appearance-geometry consistent reconstruction using only RGB input.

\subsection{Learning-based Monocular SLAM and Reconstruction}
A representative learning-based SLAM approach is~\cite{droid-slam}, which uses a recurrent neural network to predict dense optical flow increments for feed-forward tracking, significantly improving robustness in complex scenes. Recent 3D foundation models~\cite{Dust3r} rapidly generate aligned 3D point clouds from unconstrained image pairs in an end-to-end manner, marking significant progress in zero-shot reconstruction. Subsequent efforts~\cite{vggt,framevggt} extending these models to long sequences face high memory consumption and difficulty maintaining map consistency. Building on a feed-forward tracking framework and incorporating loop closure based on visual similarity with adaptive geometric alignment, our method leverages learning-based methods to achieve globally consistent reconstruction.

\section{Method}
\begin{figure*}[t]
	\centering
	\includegraphics[width=\linewidth]{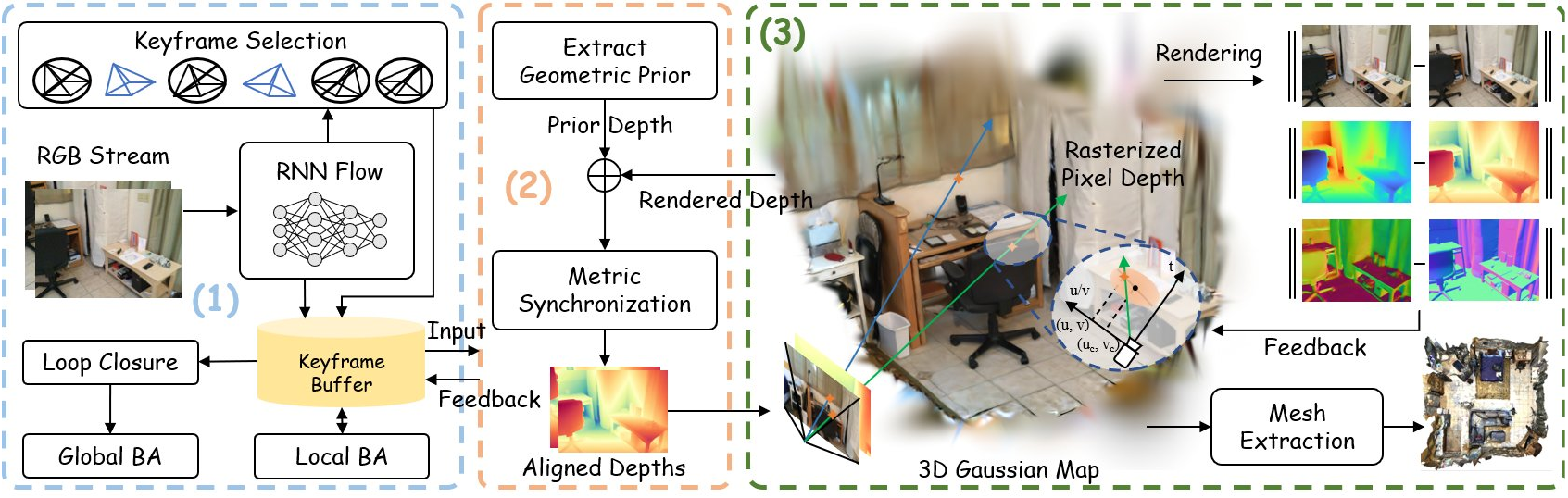}
	\caption{\textbf{Overview.} MyGO-Splat is formulated as a closed-loop geometric feedback system for RGB-only Gaussian SLAM. Given a monocular RGB video stream, a flow-based tracking frontend estimates camera poses and local geometry in real-time, while a loop-aware backend performs global BA to maintain long-term trajectory consistency. The system analytically rasterizes the optimized Gaussian map to produce multi-view consistent depth and surface normals, which are aligned with monocular geometric priors in a globally scale-consistent coordinate space and integrated into tracking optimization. Through this self-correcting feedback cycle, the system suppresses scale drift and achieves appearance-geometry consistent reconstruction without depth sensors.}
	\label{fig: overview}
	\vspace{-6pt}
\end{figure*}

The proposed MyGO-Splat aims to jointly optimize the camera trajectory $\{\mathbf{T}_i\}_{i=1}^N$ and a dense scene representation $\mathcal{G}$ (parameterized by 3D Gaussians) from an RGB stream $\{\mathbf{I}_i\}_{i=1}^N$. Unlike conventional open-loop pipelines where tracking and mapping are loosely coupled, our method establishes a bidirectional geometric feedback loop as illustrated in Fig.~\ref{fig: overview}. The system architecture is centered around three synergistic components: (1) Scale-Regulated Online Tracking (Section~\ref{sec:Tracking}), which utilizes recurrent optical flow estimation conditioned on geometric feedback; (2) Closed-Loop Geometric Feedback (Section~\ref{sec:AGA}), which serves as a metric synchronizer to project foundation-model priors into a unified coordinate frame; and (3) Appearance-Geometry Consistency Mapping via Rasterized Gaussian Geometry (Section~\ref{sec:Mapping}), which leverages the differentiable nature of Gaussian primitives to derive analytical depth and normal fields for active supervision. By continuously aligning analytically rasterized geometry with calibrated monocular priors, our framework improves the structural consistency of reconstructed surfaces.

\subsection{Online Tracking}\label{sec:Tracking}
\subsubsection{Flow-based Pose Estimation}
The tracking component employs a learning-based approach~\cite{droid-slam} that utilizes optical flow predicted by a recurrent neural network to estimate camera poses. Specifically, a keyframe graph $(\mathcal{V}, \mathcal{E})$ is constructed from the ordered input image stream $\{\mathbf{I}_i\}_{i=1}^N$, where each node in $\mathcal{V}$ represents a keyframe containing a pose $\mathbf{T} \in SE\left ( 3 \right )$ and a depth map $\mathbf{d}$. Edges in $\mathcal{E}$ are established based on dense correspondences between frames. The system determines whether to designate a frame as a new keyframe by calculating the average flow distance relative to the previous keyframe using a flow network. When the average flow exceeds a predefined threshold $d_f$, the frame is added to the keyframe sequence and buffer, while relevant geometric attributes are extracted for subsequent mapping. During the initialization phase of the system, $N$ consecutive keyframes are collected to perform initial bundle adjustment (BA) for establishing the preliminary geometric structure.

Whenever a new keyframe is inserted, the system executes dense local BA within a sliding window to jointly optimize poses $\mathbf{T}$ and depth maps $\mathbf{d}$. The objective function is constructed by minimizing the reprojection error between the predicted target points $\mathbf{p}_{ij}^{*}$ and the camera projections:
\begin{equation}
	\operatorname{E}(\mathbf{T},\mathbf{d})=\sum_{(i,j)\in\mathcal{E}}\left\|\mathbf{p}_{ij}^{*}-\Pi(\mathbf{p}_{i},\mathbf{d}_{i};\mathbf{T})\right\|_{\Sigma_{ij}}^{2}
\end{equation}
Here, a confidence weight matrix $\Sigma_{ij}$ predicted by the flow network is incorporated to mitigate the impact of occlusions or low-texture regions on the optimization, thereby enhancing the robustness of the system. The depth map $\mathbf{d}$ of the keyframes is obtained from the feedback of the subsequent mapping module. This optimization problem is solved using a damped Gauss-Newton algorithm to ensure rapid convergence during dense estimation.

\subsubsection{Loop Closure via Visual Similarity}
To ensure long-term global consistency and eliminate cumulative pose drift inherent in monocular tracking, a robust loop closure detection and global BA module is integrated. This module decouples real-time frontend tracking from global backend optimization. For loop closure detection, the system extracts global geometric features for each keyframe using~\cite{Eigenplaces} and performs efficient nearest neighbor retrieval within the historical frames using a FAISS database~\cite{faiss} to identify potential loop candidates~\cite{droid-splat}. Once a high-confidence loop is detected, corresponding cross-sequence geometric constraints are introduced into a global factor graph for global BA. This process jointly optimizes the camera poses and dense depth maps of all keyframes, achieving trajectory closure and scale correction. This provides a reliable reference for subsequent closed-loop geometric feedback and establishes a globally consistent geometric foundation for high-fidelity 3D Gaussian mapping.

\begin{table*}[h]
	\centering
	\caption{\textbf{Tracking and mapping results on the Replica dataset~\cite{replica}.} The data is derived from the mean values of eight sequences (office0-4, room0-2). "-" indicates that depth rendering or dense reconstruction is not supported.}
	\label{tab:replica}
	\renewcommand{\arraystretch}{1.2}
	\setlength{\tabcolsep}{6pt}
	\begin{tabular}{c c c c c c c c c}
		\toprule
		Cam. & Method & ATE[cm] $\downarrow$ & PSNR[dB] $\uparrow$ & SSIM $\uparrow$ & LPIPS $\downarrow$ & Acc.[cm] $\downarrow$ & Comp.[cm] $\downarrow$ & FPS[f/s] $\uparrow$ \\
		\midrule
		
		\multirow{3}{*}{\rotatebox{90}{RGB-D}}
		& SplaTAM~\cite{Splatam}     & 0.37          & 34.29          & 0.963          & 0.098          & 2.75          & 6.41          & $\sim$ 0.1            \\
		& RTG-SLAM~\cite{Rtg-slam}    & \cellcolor{orange!30}{0.25}    & 35.72          & \cellcolor{red!30}{0.980} & 0.122          & \cellcolor{yellow!30}{1.61} & {6.16} & $>5$            \\
		& GS-ICP SLAM~\cite{gs-icp} & \cellcolor{red!30}{0.19} & \cellcolor{yellow!30}{36.78} & 0.960          & \cellcolor{yellow!30}{0.054} & \cellcolor{red!30}{1.32} & \cellcolor{yellow!30}{5.91}    & $>5$            \\		
		\midrule		
		\multirow{4}{*}{\rotatebox{90}{RGB}}
		& Photo-SLAM~\cite{Photo-slam}  & 0.91          & 33.87          & 0.941          & 0.071          & -             & -             & $>5$            \\
		& MonoGS~\cite{MonoGS}      & 26.63         & 29.98          & 0.897          & 0.202          & -             & -             & $\sim$ 1.5            \\
		& Splat-SLAM~\cite{splatslam}      & 0.34         & 36.45          & 0.950          & 0.060          & 2.43             & \cellcolor{red!30}{3.64}             & $\sim$ 1.5            \\
		& SEGS-SLAM~\cite{Segs-slam}   & 0.90          & \cellcolor{orange!30}{37.79}    & \cellcolor{yellow!30}{0.964} & \cellcolor{orange!30}{0.036}    & -             & -             & $\sim$ 1.8            \\
		& \textbf{Ours}        & \cellcolor{yellow!30}{0.26} & \cellcolor{red!30}{38.33} & \cellcolor{orange!30}{0.972}    & \cellcolor{red!30}{0.032} & \cellcolor{orange!30}{1.54}    & \cellcolor{orange!30}{3.69} & $>5$            \\
		
		\bottomrule
	\end{tabular}
\end{table*}

\begin{figure*}[!ht]
	\centering
	\includegraphics[width=\linewidth]{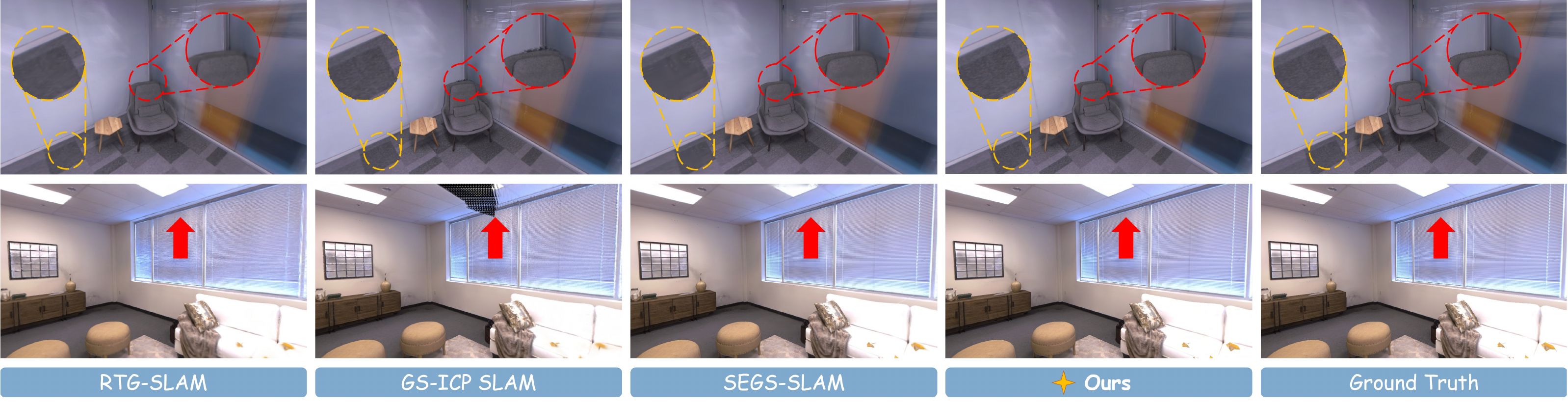}
	\caption{\textbf{Rendering results on the Replica dataset~\cite{replica}.} The arrows and the circles highlight the differences between our approach and baselines. Our method shows clearer ceiling, floor, and corner details than the baselines.}
	\label{fig:render_replica}
	\vspace{-6pt}
\end{figure*}

\subsection{Closed-Loop Geometric Feedback}
\label{sec:AGA}
While flow-based tracking provides robust pose estimation, the initial depth maps often suffer from scale drift and local distortion. To bridge scale-ambiguous monocular depth priors and metric-consistent SLAM, we propose a two-stage geometric correction framework that integrates feed-forward depth priors into the 3DGS optimization loop with a focus on spatial consistency.

\subsubsection{Geometric Prior Generation}
3D vision foundation models~\cite{vggt} estimate point clouds and poses from sparse images in a single forward pass, allowing the system to rapidly acquire high-quality geometric priors without traditional iterative optimization. Since directly incorporating 3D feed-forward models into the SLAM pipeline rarely yields more accurate pose estimation than traditional methods, the proposed approach retains only depth information to guide tracking and reconstruction.

\subsubsection{Uncertainty-Weighted Metric Synchronization}
Depth maps $D_{p}$ generated by monocular feed-forward models typically lack absolute scale and exhibit inter-frame inconsistency. The system first establishes scale alignment from global to local levels. For each keyframe, a transformation is computed to align $D_{p}$ with the rendered depth $D_{r}$ from the current 3DGS map. The optimal scale $\lambda$ and shift $\tau$ are solved by minimizing the least-squares error:
\begin{equation}
	\operatorname{E}(\lambda,\tau)=\sum_{p\in\Omega}\omega_p\left\|D_r(p)-(\lambda D_p(p)+\tau)\right\|^{2}
    \label{eq:uwms}
\end{equation}
Here, $\Omega$ denotes the set of pixels with high-confidence tracking. This procedure ensures that external geometric priors are projected into the metric coordinate space of the backend. The confidence weight $\omega_p$ is based on the photometric residuals and stability of the depth output by the 3D foundation model. 

A key novelty of this approach lies in the feedback mechanism, where refined 3DGS geometry is periodically provided to the global bundle adjustment layer. By replacing the initial noisy per-pixel depth with multi-view consistent depth rendered from 3DGS, the system gains a more stable reference to suppress the scale drift inherent in monocular SLAM, thereby establishing a self-correcting cycle between tracking and reconstruction. Different from loosely coupled tracking--mapping iterations, this feedback explicitly reuses Gaussian-rendered geometry to update the depth variables used by tracking, thereby closing the scale correction loop.

\subsection{Mapping via Rasterized Gaussian Geometry}
\label{sec:Mapping}
While standard Gaussian representations exhibit some tolerance to noise in pseudo-depth~\cite{Pseudo}, they lack constraints on fine geometric structures, which limits the further application of Gaussian maps. To address this, the system integrates a module for the computation of rasterized depth and normals. In contrast to standard 3DGS that approximates pixel depth using only Gaussian center projections, this work employs a rasterization approach to derive depth and normals. Combined with calibrated, globally consistent proxy depths, a geometric-enhanced multi-objective optimization method is designed to ensure reconstruction with appearance-geometry consistency.
\subsubsection{Rasterized Pixel Depth}
Following existing literature~\cite{3dgs}, a 3D Gaussian primitive is defined by a covariance matrix $\mathbf{\Sigma}_i$ representing its shape, which is composed of a scale $\mathbf{S}$ and rotation $\mathbf{R}$. Each Gaussian $\mathcal{G}_i=\delta_i\mathcal{N}(\mathbf{P}_i,\mathbf{\Sigma}_i)$ also includes an opacity $\delta \in [0,1]$ and color $\mathbf{c}$ represented by third-order spherical harmonics.

Inspired by~\cite{gof}, we define the intersection point between the camera ray $r$ emitted from the camera center $o$ and the 3D Gaussian $\mathcal{G}$ as the point where the Gaussian function takes its maximum value along the ray direction. However, instead of using the time-consuming ray tracing method to estimate the depth of the Gaussian elements as in~\cite{gof,FGO-SLAM}, we adopt a rasterization approach. Under a local affine projection, the 3D Gaussian is transformed into a ray space where the ray direction aligns with the z-axis. Following~\cite{radegs}, it is derived that the set of intersection points between a Gaussian primitive and a bundle of rays is geometrically coplanar in this space. This property enables the efficient rasterization of the precise depth $d$ for each pixel $x=(u, v)$ through a plane equation:

\begin{equation}
d=z_o+\hat{\mathbf{p}}^{\top}\begin{bmatrix}u_o-u\\v_o-v\end{bmatrix}
\end{equation}
where $z_o$ denotes the depth of the center of the Gaussian, $(u_o, v_o)$ represents the pixel coordinates of the center of the Gaussian, and $\hat{\mathbf{p}} \in \mathbb{R}^2$ is a coefficient vector determined by the covariance of the Gaussian and camera intrinsics. This formulation enables Gaussian primitives to represent pixel-level depth variations, which significantly improves the continuity and accuracy of the resulting depth maps.

\subsubsection{Surface Analytical Normal}
The planar characteristics of the intersection points in ray space also allow for the analytical computation of surface normals directly from the parameters of the Gaussians. The plane equation in ray space defines the normal:
\begin{equation}
	%\mathbf{n}_{\mathrm{ray}}=-(\frac{t_{o}}{z_{o}}\mathbf{\hat{p}}\quad1)^{\mathrm{T}}
	\mathbf{n}_{\mathrm{ray}}=-\left[\frac{t_o}{z_o}\hat{\mathbf{p}}^{\top},\,1\right]^{\top}
\end{equation}
which is subsequently transformed back to the camera coordinate system using the transpose of the Jacobian matrix $\mathbf{J}$ of the local affine transformation:
\begin{equation}
	\mathbf{n}_{\mathrm{cam}}=\mathbf{J}^\mathrm{T}\mathbf{n}_{\mathrm{ray}}
\end{equation}
This analytical normal $\mathbf{n}_{\mathrm{cam}}$ provides geometric supervision for the SLAM system, particularly in textureless regions where photometric loss typically fails.

\begin{table}[!tpb]
	\centering
	\caption{\textbf{Tracking and mapping results on the TUM RGB-D dataset~\cite{tum}.} The data is derived from the mean values of three sequences (fr1\_desk, fr2\_xyz, fr3\_office).}
	\label{tab:tum}
	\renewcommand{\arraystretch}{1.2}
	\setlength{\tabcolsep}{3pt}
	\begin{tabular}{c c c c c c}
		\toprule
		Cam. & Method & ATE[cm] $\downarrow$ & PSNR[dB] $\uparrow$ & SSIM $\uparrow$ & LPIPS $\downarrow$ \\
		\midrule
		
		\multirow{4}{*}{\rotatebox{90}{RGB-D}}
		& SplaTAM~\cite{Splatam}    & 3.41            & 22.45            & \cellcolor{red!30}{0.839}  & 0.214 \\
		& MonoGS~\cite{MonoGS}     & 1.55            & {24.02}   & 0.785           & 0.240 \\
		& Photo-SLAM~\cite{Photo-slam} & \cellcolor{yellow!30}{1.38}   & 21.51            & 0.748           & \cellcolor{yellow!30}{0.211} \\
		& RTG-SLAM~\cite{Rtg-slam}   & \cellcolor{red!30}{1.09}   & 17.29            & 0.595           & 0.466 \\
		
		\midrule
		
		\multirow{4}{*}{\rotatebox{90}{RGB}}
		& MonoGS~\cite{MonoGS}     & 4.53            & 21.88            & 0.733           & 0.342 \\
		& Photo-SLAM~\cite{Photo-slam} & 1.47            & 19.02            & 0.689           & 0.433 \\
		& SEGS-SLAM~\cite{Segs-slam}  & 1.67            & \cellcolor{orange!30}{24.75}& \cellcolor{yellow!30}{0.801}  & \cellcolor{orange!30}{0.129} \\
		& \textbf{Ours}       & \cellcolor{orange!30}{1.14}& \cellcolor{red!30}{25.01}   & \cellcolor{orange!30}{0.813} & \cellcolor{red!30}{0.125} \\
		
		\bottomrule
	\end{tabular}
    \vspace{-4pt}
\end{table}

\subsubsection{Geometric-Enhanced Multi-Objective Optimization}
The objective is designed to jointly enforce appearance fidelity and geometric coherence under a closed-loop optimization framework. Since these extensions remain compatible with the general Gaussian representation, an appearance optimization function is naturally constructed:
\begin{equation}
	\mathcal{L}_{\mathrm{rgb}}=\sum_{k\in\mathcal{K}}\left|\mathbf{I}_k^{\prime}-\mathbf{I}_k\right|
\end{equation}
where $\mathcal{K}$ represents the set of keyframes. The rendered color $\mathbf{I}^{\prime}$ is computed via alpha-blending along each ray from near to far:
\begin{equation}
	C\left(p\right)=\sum_{i=1}^{N}c_i\omega_i\prod_{j=1}^{i-1}\left(1-\omega_j\right)
\end{equation} 
Here, $N$ is the number of Gaussians along the ray, and $\omega_i$ is the pixel translucency determined by the $i$-th Gaussian and pixel position $p$. 
Similarly, the rasterized pixel depth is obtained:
\begin{equation}
	D_r\left(p\right)=\sum_{i=1}^{N}d_i\omega_i\prod_{j=1}^{i-1}\left(1-\omega_j\right)
\end{equation} 

By calculating the loss between rendered depth $\mathbf{D}^{\prime}$ and the aligned proxy depth $\mathbf{D}$ (calculated from Equation~\ref{eq:uwms} based on the depth $D_p(p)$ of each pixel after synchronization), the depth optimization function is established:
\begin{equation}
	\mathcal{L}_{\mathrm{d}}=\sum_{k\in\mathcal{K}}\left|\mathbf{D}_k^{\prime}-\mathbf{D}_k\right|
\end{equation}

Furthermore, two additional regularization terms are introduced to refine the geometric structure of the reconstructed scene:
\begin{equation}
	\mathcal{L}_{\mathrm{dis}}=\sum_{i,j}\delta_{i}\delta_{j}\left|d_{i}-d_{j}\right|, \ \mathcal{L}_{n}=\sum_{k\in\mathcal{K}}\left|1-\mathbf{N}_{k}^{\prime T}\cdot\mathbf{N}_{k}\right|
\end{equation}
Here, the term $\mathcal{L}_{\mathrm{dis}}$ aims to minimize the depth discrepancy between different Gaussians along the same ray to reduce artifacts, while $\mathcal{L}_{n}$ represents the consistency error between the analytical normals $\mathbf{N}_{k}^{\prime}$ derived from the Gaussian scene and the normals $\mathbf{N}_{k}$ generated from the prior depth. The complete optimization function is formulated as follows:
\begin{equation}
\mathcal{L}=\lambda_{\mathrm{rgb}}\mathcal{L}_{\mathrm{rgb}}+\lambda_\mathrm{d}\mathcal{L}_{\mathrm{d}}+\lambda_{\mathrm{dis}}\mathcal{L}_{\mathrm{dis}}+\lambda_\mathrm{n}\mathcal{L}_{\mathrm{n}}
\label{eq:final_loss}
\end{equation}
where $\lambda_\mathrm{rgb}$, $\lambda_\mathrm{d}$, $\lambda_\mathrm{dis}$, and $\lambda_\mathrm{n}$ are hyperparameters that balance the contribution of each loss term.

\section{Experiment}
\subsection{Experiment Setup}

\begin{table}[!tpb]
	\centering
	\caption{\textbf{Tracking and mapping results on the ScanNet dataset~\cite{scannet}.} The data is derived from the mean values of six sequences (0000, 0059, 0106, 0169, 0181, 0207).}
	\label{tab:scannet}
	\renewcommand{\arraystretch}{1.2}
	\setlength{\tabcolsep}{3pt}
	\begin{tabular}{c c c c c c}
		\toprule
		Cam. & Method & ATE[cm] $\downarrow$ & PSNR[dB] $\uparrow$ & SSIM $\uparrow$ & LPIPS $\downarrow$ \\
		\midrule
		
		\multirow{4}{*}{\rotatebox{90}{RGB-D}}
		& SplaTAM~\cite{Splatam}    & 11.71 & {20.55} & 0.725 & {0.306} \\
		& MonoGS~\cite{MonoGS}     & 12.80 & \cellcolor{yellow!30}{20.62} & {0.756} & \cellcolor{yellow!30}{0.298} \\
		& Photo-SLAM~\cite{Photo-slam} & {10.56} & 19.62 & \cellcolor{yellow!30}{0.773} & 0.349 \\
		& RTG-SLAM~\cite{Rtg-slam}   & 11.25 & 16.85 & 0.750 & 0.488 \\
		
		\midrule
		
		\multirow{3}{*}{\rotatebox{90}{RGB}}
		& MonoGS~\cite{MonoGS}     & 105.6          & 18.75 & 0.693 & 0.355 \\
		& Splat-SLAM~\cite{splatslam}     & \cellcolor{orange!30}{7.59}          & \cellcolor{red!30}{29.25} & \cellcolor{orange!30}{0.852} & \cellcolor{red!30}{0.193} \\
		& SEGS-SLAM~\cite{Segs-slam}  & \cellcolor{yellow!30}{8.59}           & 19.58 & 0.705 & 0.348 \\
		& \textbf{Ours}       & \cellcolor{red!30}{7.22}  & \cellcolor{orange!30}{29.09} & \cellcolor{red!30}{0.878} & \cellcolor{orange!30}{0.232} \\
		
		\bottomrule
	\end{tabular}
    \vspace{-4pt}
\end{table}

\begin{figure*}[h]
	\centering
	\includegraphics[width=\linewidth]{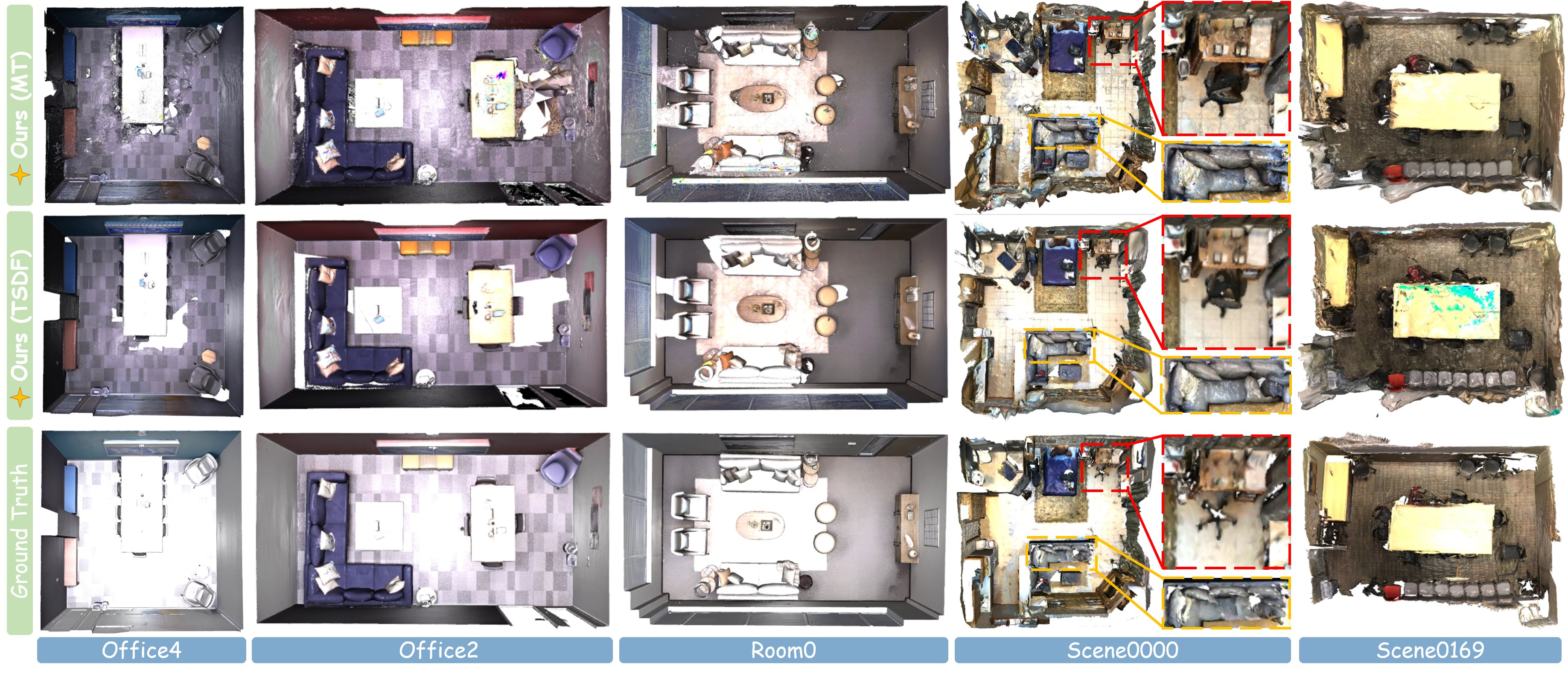}
	\caption{\textbf{Mesh reconstruction results in the Replica dataset~\cite{replica} and the ScanNet dataset~\cite{scannet}.} Our method can utilize the common mesh extraction techniques and effectively restore the geometric structure and details.}
	\label{fig:mesh}
	\vspace{-6pt}
\end{figure*}

\begin{figure}[!tpb]
	\centering
	\includegraphics[width=\linewidth]{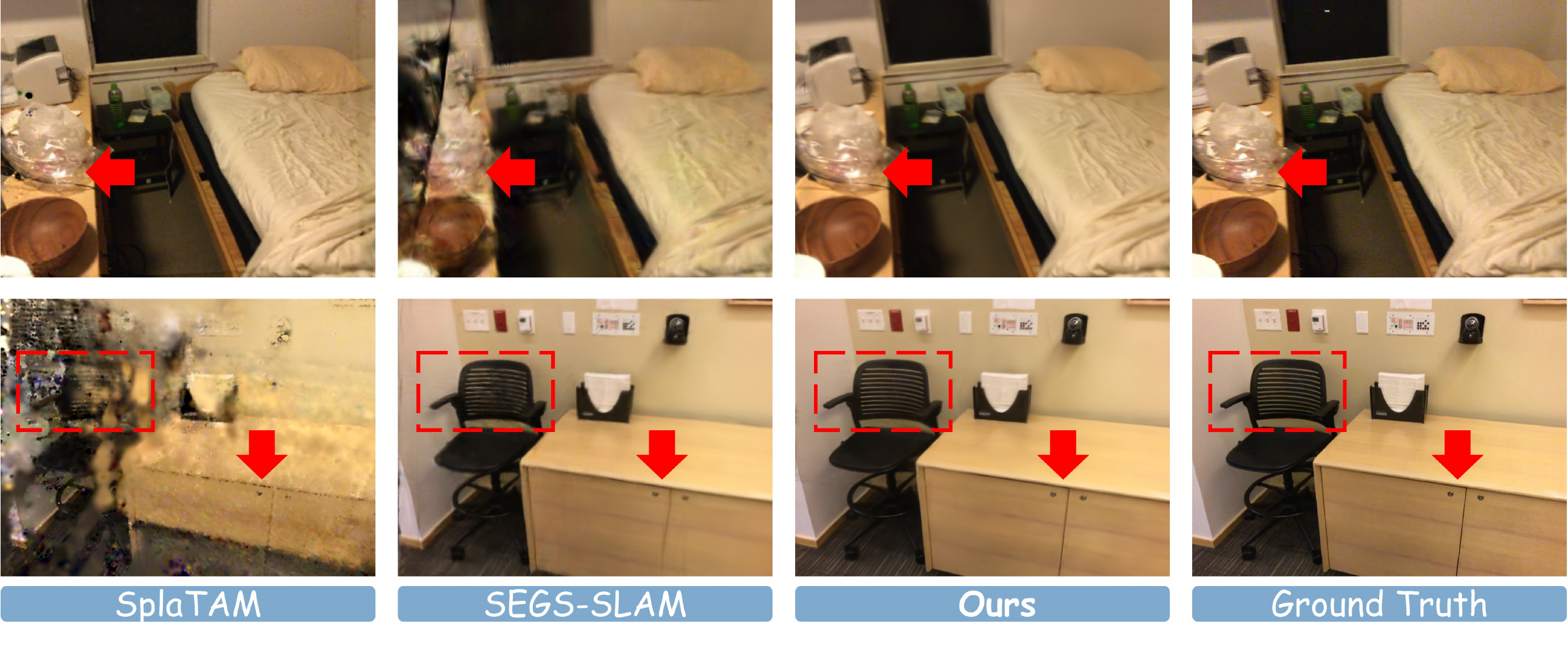}
	\caption{\textbf{Rendering results on the ScanNet dataset~\cite{scannet}.} The red arrows and the boxes highlight the differences between our approach and baselines.}
	\label{fig:render_scannet}
	\vspace{-6pt}
\end{figure}

\subsubsection{Implementation Details}
The proposed system is implemented using Python 3 and CUDA. Evaluations are conducted on a workstation equipped with an NVIDIA A100 GPU (40 GB VRAM) and an AMD EPYC 7542 CPU. The system utilizes default parameters from~\cite{3dgs,droid-slam}. Specifically, the filtering mechanism follows~\cite{Mip-splatting}, while the densification strategy is based on the approach proposed in~\cite{gof}. An Adam optimizer is employed to refine the parameters of the Gaussian primitives. The weighting hyperparameters $\lambda_{rgb}$, $\lambda_{d}$, $\lambda_{dis}$, and $\lambda_{n}$ in Equation~\ref{eq:final_loss} are set to 1, 0.2, 1000, and 0.1 (0.5 for ScanNet).

\subsubsection{Datasets and Evaluation Metrics}
Performance is evaluated on real-world datasets, including TUM RGB-D~\cite{tum} and ScanNet~\cite{scannet}, as well as the synthetic Replica dataset~\cite{replica}, which provides accurate ground-truth data. The selection of sequences follows the configurations established in~\cite{Nice-slam}. To quantitatively evaluate the performance of the system, standard metrics are utilized in accordance with NICE-SLAM~\cite{Nice-slam} and SplaTAM~\cite{Splatam}. Pose estimation is evaluated using the average root mean square error of the absolute trajectory error (ATE RMSE)~\cite{tum}. Appearance reconstruction is measured by PSNR, SSIM, and LPIPS, while the quality of geometric reconstruction is assessed through accuracy (Acc.) and completion (Comp.).

\subsubsection{Baseline Comparisons}
The proposed method is compared against state-of-the-art (SOTA) approaches based on 3DGS, including  Photo-SLAM~\cite{Photo-slam}, MonoGS~\cite{MonoGS}, SplaTAM~\cite{Splatam}, RTG-SLAM~\cite{Rtg-slam}, GS-ICP SLAM~\cite{gs-icp}, Splat-SLAM~\cite{splatslam}, and SEGS-SLAM~\cite{Segs-slam}. All baseline methods are executed using their official implementations, and the results are reported as the average of five independent runs. The \colorbox{red!30}{best}, \colorbox{orange!30}{second}, and \colorbox{yellow!30}{third} results are highlighted in the experimental tables. For fairness, all RGB-only baselines are evaluated with their official RGB input settings, and RGB-D methods are reported separately because they use sensor depth. MyGO-Splat uses VGGT~\cite{vggt} only as a pretrained monocular geometric prior and does not use ground-truth depth from the evaluation datasets.

\subsection{Camera Pose Estimation}
Table~\ref{tab:replica},~\ref{tab:tum}, and~\ref{tab:scannet} show the camera tracking accuracy of the proposed system on the real-world TUM RGB-D, ScanNet, and synthetic Replica datasets. MyGO-Splat operates online and delivers competitive pose estimation by correcting monocular scale drift through a mapping-to-tracking feedback loop with consistent rendered depth, achieving performance near RGB-D methods. Visual similarity-based loop closures further support overall tracking accuracy. The large ATE gap on ScanNet mainly reflects the difficulty of maintaining global scale in challenging RGB-only sequences, where feedback from rendered geometry provides a stable scale reference.

\subsection{Appearance Rendering Quality}
Table~\ref{tab:replica},~\ref{tab:tum}, and~\ref{tab:scannet} compare our rendering quality with recent SOTA approaches. MyGO-Splat achieves the highest rendering quality among RGB-only methods, performing comparably to SOTA RGB-D methods. Fig.~\ref{fig:render_replica} and~\ref{fig:render_scannet} show that the proposed method better preserves floor textures, wall corners, ceiling edges, and high-frequency textures. This improvement stems from precise Gaussian primitive distribution constraints. Collaborative geometric-enhanced multi-objective optimization forces Gaussian primitives to distribute along actual physical surfaces, eliminating visual artifacts while producing sharper edges and higher texture alignment.

\subsection{Geometric Reconstruction Accuracy}
Table~\ref{tab:replica} quantitatively evaluates Replica geometric reconstruction, where the proposed method leads in accuracy and completion. Fig.~\ref{fig:mesh} illustrates meshes extracted by marching tetrahedra (MT) and TSDF fusion. Replica ground truth is software-synthesized, while ScanNet ground truth is derived from traditional dense visual SLAM. Our method produces accurate geometric restoration on Replica and visually more complete fine structures on ScanNet, such as chairs and tables. Since ScanNet reconstructions from traditional fusion may contain sensor noise or holes at thin structures and edges, rasterized Gaussians generate continuous depth maps for more complete scene representation.

\begin{table}[!tpb]
	\centering
	\caption{\textbf{Ablation study on the key components.} The data is derived from the mean values of eight sequences in the Replica dataset~\cite{replica}.}
	\label{tab:ablation}
	\renewcommand{\arraystretch}{1.2}
	\setlength{\tabcolsep}{4pt}
	\begin{tabular}{l c c c c}
		\toprule
		\multirow{2}{*}{Method} 
		& ATE & PSNR & Acc. & Comp. \\
		& [cm] $\downarrow$ 
		& [dB] $\uparrow$ 
		& [cm] $\downarrow$ 
		& [cm] $\downarrow$ \\
		\midrule
		
		w/o Loop Closure                 & 0.45          & 37.95          & \cellcolor{orange!30}{1.82}    & \cellcolor{orange!30}{4.12} \\
		w/o CLG Feedback & 0.58          & 35.12          & 3.86                & 5.24 \\
		w/o GEMO Optimization     & \cellcolor{orange!30}{0.29} & \cellcolor{red!30}{38.45} & 5.42            & 9.85 \\
		Full Implementation              & \cellcolor{red!30}{0.26} & \cellcolor{orange!30}{38.33} & \cellcolor{red!30}{1.54} & \cellcolor{red!30}{3.69} \\
		
		\bottomrule
	\end{tabular}
\end{table}

\subsection{Ablation Study}
\begin{figure}[!tpb]
	\centering
	\begin{subfigure}[b]{0.49\linewidth}
		\centering
		\includegraphics[width=\linewidth]{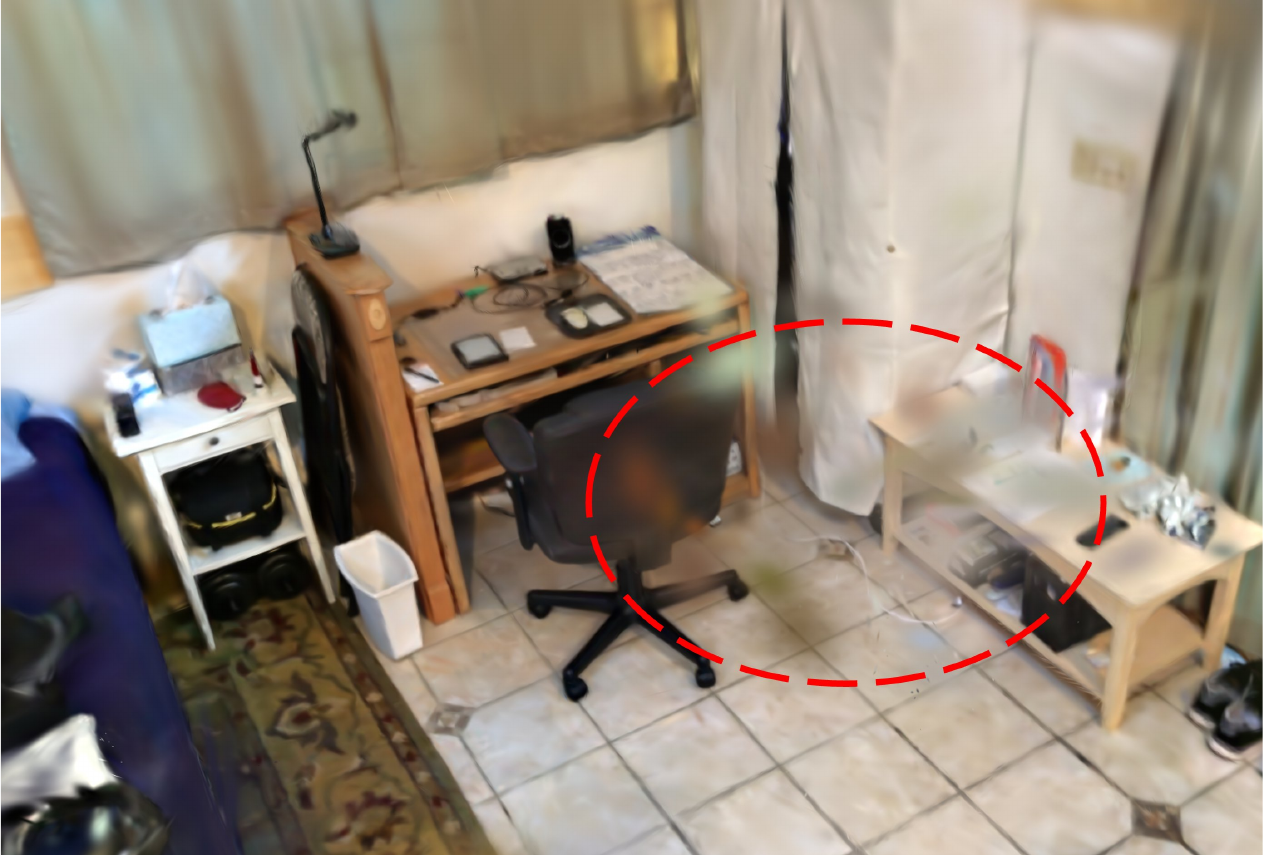}
		\caption{w/o CLG Feedback}
	\end{subfigure}
	\hfill
	\begin{subfigure}[b]{0.49\linewidth}
		\centering
		\includegraphics[width=\linewidth]{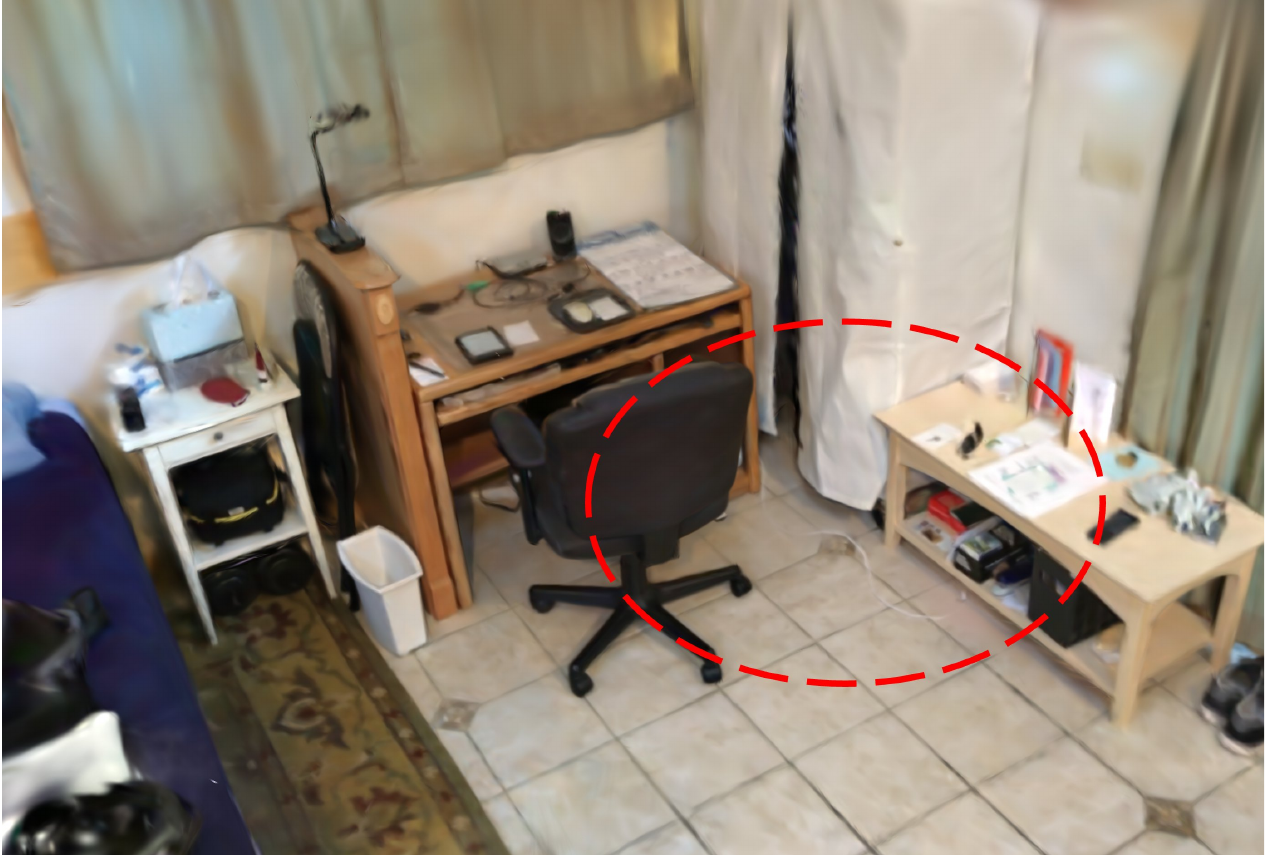}
		\caption{w CLG Feedback}
	\end{subfigure}
	\caption{\textbf{Ablation of closed-loop geometric feedback.} The red dotted circle highlights the improvement in reconstruction quality achieved by using the CLG Feedback method. The scene is from scene0000 sequence of ScanNet dataset~\cite{scannet}.}
	\label{fig:aga}
	\vspace{-4pt}
\end{figure}

A series of ablation experiments evaluates the contributions of loop closure via visual similarity, closed-loop geometric (CLG) feedback, and geometric-enhanced multi-objective (GEMO) optimization. Quantitative results are reported in Table~\ref{tab:ablation}, while qualitative visual differences are illustrated in Fig.~\ref{fig:aga} and~\ref{fig:reg}.
%The experiments demonstrate that each module is essential for maintaining global consistency and achieving high-fidelity reconstruction.

\subsubsection{Loop Closure via Visual Similarity}
As shown in Table~\ref{tab:ablation}, removing loop closure increases the ATE RMSE from 0.26 cm to 0.45 cm, indicating that cumulative geometric drift in monocular tracking cannot be effectively corrected without global constraints. By introducing cross-sequence constraints and global bundle adjustment, loop closure maintains trajectory consistency and high localization accuracy during long-sequence operation, while having only a minor impact on local rendering metrics such as PSNR.

\subsubsection{Closed-Loop Geometric Feedback}
Removing CLG feedback increases the ATE to 0.58 cm and drops the PSNR to 35.12 dB. Since significant scale and shift discrepancies exist between the prior depth generated by 3D vision foundation models and the metric space of the SLAM system, a lack of alignment causes the depth feedback mechanism to fail and induces severe scale drift. Fig.~\ref{fig:aga} illustrates the impact of CLG Feedback on the reconstructed Gaussian maps: without geometric alignment, the maps exhibit prominent floaters and ghosting effects, whereas CLG Feedback projects prior geometry into the metric space and allows Gaussian primitives to cluster tightly on physical surfaces.

\subsubsection{Geometric-Enhanced Multi-Objective Optimization}
\begin{figure}[!tpb]
	\centering
	\begin{subfigure}[b]{0.49\linewidth}
		\centering
		\includegraphics[width=\linewidth]{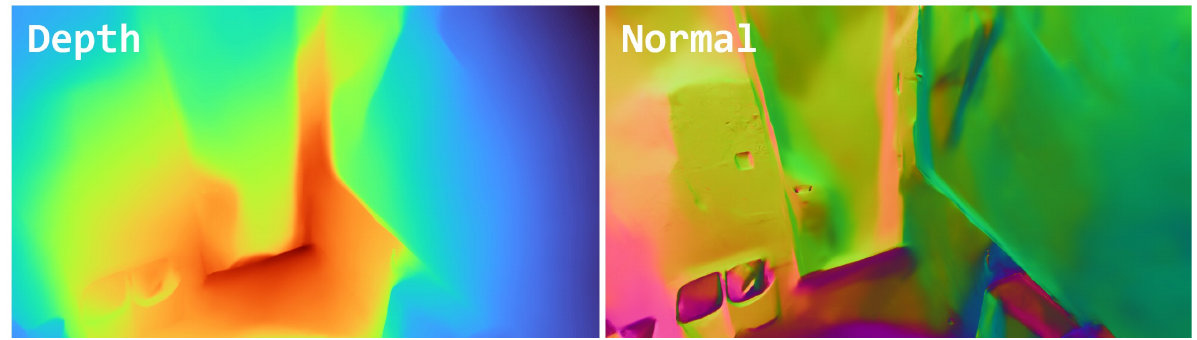}
		\caption{w/o GEMO Regularization}
	\end{subfigure}
	\hfill
	\begin{subfigure}[b]{0.49\linewidth}
		\centering
		\includegraphics[width=\linewidth]{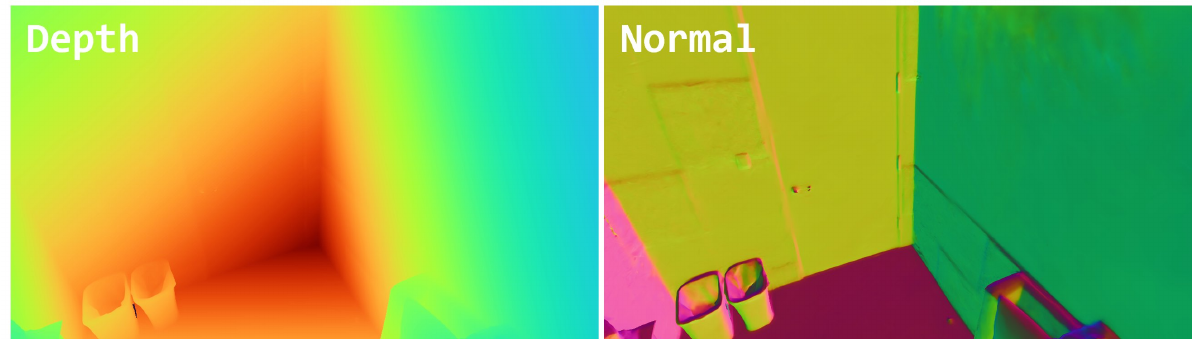}
		\caption{w GEMO Regularization}
	\end{subfigure}
	\caption{\textbf{Ablation of geometric-enhanced multi-objective optimization.} Our comprehensive approach has demonstrated a significant improvement in the geometric structure. The scene is from office1 sequence of Replica dataset~\cite{replica}.}
	\label{fig:reg}
	\vspace{-6pt}
\end{figure}

The geometric-enhanced multi-objective optimization module improves the topological quality of reconstructed surfaces, although geometric constraints slightly reduce appearance metrics. Removing the regularization degrades mesh accuracy from 1.54 cm to 5.42 cm and increases completion from 3.69 cm to 9.85 cm. Fig.~\ref{fig:reg} compares rendered depth and normal maps: without regularization, the Gaussian distribution appears distorted and fails to form a continuous surface, whereas geometric regularization produces smooth and detail-rich edges for high-quality mesh extraction.

\subsection{Demonstration}
MyGO-Splat is further tested on a real-world outdoor video sequence captured by a drone using only monocular RGB input, without depth sensors or ground-truth depth. Since metric ground truth is unavailable for this outdoor sequence, we use it as a qualitative demonstration of online operation rather than a quantitative benchmark. The system performs monocular pose estimation and Gaussian map construction at 5 fps while extracting continuous meshes, as illustrated in Fig.~\ref{fig:abstract} with multimodal rendered data and high-quality mesh reconstructions. 

\section{Conclusion}
We present a closed-loop geometric feedback monocular SLAM system based on 3DGS that addresses scale drift and limited geometric fidelity in RGB-only reconstruction. By coupling feed-forward pose inference with global BA and projecting pseudo-depth priors into a globally scale-consistent Gaussian space, the system achieves robust and accurate trajectory estimation while maintaining real-time performance. Joint optimization over rasterized depth and normals enforces appearance-geometry consistency. Experiments on standard datasets demonstrate state-of-the-art localization and reconstruction performance, enabling direct extraction of high-quality meshes from monocular video.

%We propose a multi-source depth-guided monocular SLAM system based on 3DGS, which addresses the persistent challenges of scale drift and limited geometric fidelity in RGB-only scene reconstruction. By integrating feed-forward pose inference with a globally consistent loop closure detection mechanism, the system achieves robust and high-precision camera trajectory estimation while maintaining real-time performance. The adaptive geometric alignment module transforms pseudo-depth priors into precise constraints within a globally scale-consistent space, while a collaborative optimization strategy based on rasterized depth and normals significantly enhances the consistency between the appearance and geometry of the scene reconstruction. Experimental results on several standard datasets demonstrate that the proposed method achieves state-of-the-art performance in both localization accuracy and reconstruction quality, facilitating the direct extraction of high-quality meshes from monocular video streams.

%\section*{ACKNOWLEDGMENT}

%\newpage
%\bibliographystyle{unsrt}   
\bibliographystyle{IEEEtran}
\bibliography{reference}

@String(CVPR= {IEEE Conf. Comput. Vis. Pattern Recog.})

@String(ICCV= {Int. Conf. Comput. Vis.})

@String(ECCV= {Eur. Conf. Comput. Vis.})

@String(TOG= {ACM Trans. Graph.})

@String(CVPR  = {CVPR})

@String(ICCV  = {ICCV})

@String(ECCV  = {ECCV})

@String(TOG   = {ACM TOG})

@inproceedings{Scannet,
	author = {A. Dai and A. X. Chang and M. Savva and M. Halber and T. Funkhouser and M. Nießner},
	title = {Scannet: Richly-annotated 3d reconstructions of indoor scenes},
	booktitle = {Proceedings of the IEEE/CVF Conference on Computer Vision and Pattern Recognition (CVPR)},
	year = {2017},
	pages = {5828-5839},
	type = {Conference Proceedings}
}

@inproceedings{Photo-slam,
	author = {H. Huang and L. Li and H. Cheng and S.-K. Yeung},
	title = {Photo-slam: Real-time simultaneous localization and photorealistic mapping for monocular stereo and rgb-d cameras},
	booktitle = {Proceedings of the IEEE/CVF Conference on Computer Vision and Pattern Recognition (CVPR)},
	year = {2024},
	pages = {21584-21593},
	type = {Conference Proceedings}
}

@inproceedings{Splatam,
	author = {N. Keetha and J. Karhade and K. M. Jatavallabhula and G. Yang and S. Scherer and D. Ramanan and J. Luiten},
	title = {Splatam: Splat track \& map 3d gaussians for dense rgb-d slam},
	booktitle = {Proceedings of the IEEE/CVF Conference on Computer Vision and Pattern Recognition (CVPR)},
	year = {2024},
	pages = {21357-21366},
	type = {Conference Proceedings}
}

@article{3dgs,
	author = {B. Kerbl and G. Kopanas and T. Leimkühler and G. Drettakis},
	title = {3d gaussian splatting for real-time radiance field rendering},
	journal = {ACM Transactions on Graphics},
	volume = {42},
	number = {4},
	pages = {1-14},
	year = {2023},
	type = {Journal Article}
}

@inproceedings{MonoGS,
	author = {H. Matsuki and R. Murai and P. H. Kelly and A. J. Davison},
	title = {Gaussian splatting slam},
	booktitle = {Proceedings of the IEEE/CVF Conference on Computer Vision and Pattern Recognition (CVPR)},
	year = {2024},
	pages = {18039-18048},
	type = {Conference Proceedings}
}

@article{Nerf,
	author = {B. Mildenhall and P. P. Srinivasan and M. Tancik and J. T. Barron and R. Ramamoorthi and R. Ng},
	title = {Nerf: Representing scenes as neural radiance fields for view synthesis},
	journal = {Communications of the ACM},
	volume = {65},
	number = {1},
	pages = {99-106},
	year = {2022},
	type = {Journal Article}
}

@inproceedings{Rtg-slam,
	author = {Z. Peng and T. Shao and Y. Liu and J. Zhou and Y. Yang and J. Wang and K. Zhou},
	title = {Rtg-slam: Real-time 3d reconstruction at scale using gaussian splatting},
	booktitle = {ACM SIGGRAPH},
	year = {2024},
	pages = {1-11},
	type = {Conference Proceedings}
}

@article{replica,
	author = {J. Straub and T. Whelan and L. Ma and Y. Chen and E. Wijmans and S. Green and J. J. Engel and R. Mur-Artal and C. Ren and S. Verma},
	title = {The replica dataset: A digital replica of indoor spaces},
	journal = {arXiv preprint arXiv:1906.05797},
	year = {2019},
	type = {Journal Article}
}

@inproceedings{tum,
	author = {J. Sturm and N. Engelhard and F. Endres and W. Burgard and D. Cremers},
	title = {A benchmark for the evaluation of rgb-d slam systems},
	booktitle = {2012 IEEE/RSJ International Conference on Intelligent Robots and Systems (IROS)},
	pages = {573-580},
	year = {2012},
	type = {Conference Proceedings}
}

@inproceedings{Imap,
	author = {E. Sucar and S. K. Liu and J. Ortiz and A. J. Davison},
	title = {Imap: Implicit mapping and positioning in real-time},
	booktitle = {Proceedings of the IEEE/CVF International Conference on Computer Vision (ICCV)},
	pages = {6209-6218},
	year = {2021},
	type = {Conference Proceedings}
}

@inproceedings{Nice-slam,
	author = {Z. Zhu and S. Peng and V. Larsson and W. Xu and H. Bao and Z. Cui and M. R. Oswald and M. Pollefeys},
	title = {Nice-slam: Neural implicit scalable encoding for slam},
	booktitle = {Proceedings of the IEEE/CVF Conference on Computer Vision and Pattern Recognition (CVPR)},
	year = {2022},
	pages = {12776-12786},
	type = {Conference Proceedings}
}

@article{HI-SLAM2,
	title={Hi-slam2: Geometry-aware gaussian slam for fast monocular scene reconstruction},
	author={Zhang, Wei and Cheng, Qing and Skuddis, David and Zeller, Niclas and Cremers, Daniel and Haala, Norbert},
	journal={IEEE Transactions on Robotics},
	volume={41},
	pages={6478--6493},
	year={2025},
	publisher={IEEE}
}

@inproceedings{FGO-SLAM,
  author={Zhu, Fan and Zhao, Yifan and Chen, Ziyu and Yu, Biao and Zhu, Hui},
  booktitle={2025 IEEE International Conference on Robotics and Automation (ICRA)}, 
  title={FGO-SLAM: Enhancing Gaussian SLAM with Globally Consistent Opacity Radiance Field}, 
  year={2025},
  pages={11075-11081},
  organization={IEEE}
}

@inproceedings{gs-icp,
  title={Rgbd gs-icp slam},
  author={Ha, Seongbo and Yeon, Jiung and Yu, Hyeonwoo},
  booktitle={Proceedings of the European Conference on Computer Vision (ECCV)},
  pages={180--197},
  year={2024},
  organization={Springer}
}

@inproceedings{GARAD-SLAM,
  author={Li, Mingrui and Chen, Weijian and Cheng, Na and Xu, Jingyuan and Li, Dong and Wang, Hongyu},
  booktitle={2025 IEEE International Conference on Robotics and Automation (ICRA)}, 
  title={GARAD-SLAM: 3D Gaussian Splatting for Real-Time Anti Dynamic SLAM}, 
  year={2025},
  pages={11047-11053},
  organization={IEEE}
}

@article{Bundlefusion,
  title={Bundlefusion: Real-time globally consistent 3d reconstruction using on-the-fly surface reintegration},
  author={Dai, Angela and Nie{\ss}ner, Matthias and Zollh{\"o}fer, Michael and Izadi, Shahram and Theobalt, Christian},
  journal={ACM Transactions on Graphics},
  volume={36},
  number={4},
  pages={1},
  year={2017},
  publisher={ACM New York, NY, USA}
}

@inproceedings{SLAM-X,
  title={SLAM-X: Generalizable dynamic removal for NeRF and Gaussian splatting SLAM},
  author={Li, Mingrui and Li, Dong and Hu, Sijia and Wang, Kangxu and Zhao, Zhenjun and Wang, Hongyu},
  booktitle={Proceedings of the 33rd ACM International Conference on Multimedia},
  pages={1132--1140},
  year={2025}
}

@article{advances3dcv,
	title={Advances in Global Solvers for 3D Vision},
	author={Zhao, Zhenjun and Yang, Heng and Liao, Bangyan and Zeng, Yingping and Yan, Shaocheng and Gu, Yingdong and Liu, Peidong and Zhou, Yi and Li, Haoang and Civera, Javier},
	journal={arXiv preprint arXiv:2602.14662},
	year={2026}
}

@inproceedings{ULF-Loc,
	title={ULF-Loc: Unbiased Landmark Feature for Robust Visual Localization with 3D Gaussian Splatting},
	author={Gu, Yingdong and Yan, Shaocheng and Zhao, Zhenjun and Kou, Yuan and Luo, Jianxin and Shi, Pengcheng and Li, Jiayuan},
	booktitle={Proceedings of the IEEE/CVF Conference on Computer Vision and Pattern Recognition},
	year={2026}
}

@article{PanoImager,
	title={PanoImager: Geometry-Guided Novel View Synthesis and Reconstruction from Sparse Panoramic Views}, 
	author={Xu, Zhisong and Oishi, Takeshi},
	journal={arXiv preprint arXiv:2606.27071},
	year={2026}
}

@misc{Slamhandbook,
	title={Slam handbook: From localization and mapping to spatial intelligence},
	author={Carlone, Luca and Kim, Ayoung and Barfoot, Timothy and Cremers, Daniel and Dellaert, Frank},
	year={2025},
	publisher={Cambridge University Press}
}

@article{dygs,
	title={DyGS-SLAM: Realistic Map Reconstruction in Dynamic Scenes Based on Double-Constrained Visual SLAM},
	author={Zhu, Fan and Zhao, Yifan and Chen, Ziyu and Jiang, Chunmao and Zhu, Hui and Hu, Xiaoxi},
	journal={Remote Sensing},
	volume={17},
	number={4},
	pages={625},
	year={2025},
	publisher={MDPI}
}

@inproceedings{Pseudo,
	title={Pseudo Depth Meets Gaussian: A Feed-forward RGB SLAM Baseline},
	author={Zhao, Linqing and Xu, Xiuwei and Wang, Yirui and Wang, Hao and Zheng, Wenzhao and Tang, Yansong and Yan, Haibin and Lu, Jiwen},
	booktitle={2025 IEEE/RSJ International Conference on Intelligent Robots and Systems (IROS)},
	pages={8142--8149},
	year={2025},
	organization={IEEE}
}

@inproceedings{Dust3r,
	title={Dust3r: Geometric 3d vision made easy},
	author={Wang, Shuzhe and Leroy, Vincent and Cabon, Yohann and Chidlovskii, Boris and Revaud, Jerome},
	booktitle={Proceedings of the Computer Vision and Pattern Recognition Conference (CVPR)},
	pages={20697--20709},
	year={2024}
}

@article{droid-slam,
	title={Droid-slam: Deep visual slam for monocular, stereo, and rgb-d cameras},
	author={Teed, Zachary and Deng, Jia},
	journal={Advances in Neural Information Processing Systems},
	volume={34},
	pages={16558--16569},
	year={2021}
}

@inproceedings{Segs-slam,
	title={Segs-slam: Structure-enhanced 3d gaussian splatting slam with appearance embedding},
	author={Wen, Tianci and Liu, Zhiang and Fang, Yongchun},
	booktitle={Proceedings of the IEEE/CVF International Conference on Computer Vision (ICCV)},
	pages={28103--28113},
	year={2025}
}

@inproceedings{vggt,
	title={Vggt: Visual geometry grounded transformer},
	author={Wang, Jianyuan and Chen, Minghao and Karaev, Nikita and Vedaldi, Andrea and Rupprecht, Christian and Novotny, David},
	booktitle={Proceedings of the Computer Vision and Pattern Recognition Conference (CVPR)},
	pages={5294--5306},
	year={2025}
}

@inproceedings{sage,
	title={{SAGE}: Spatial-visual Adaptive Graph Exploration for Efficient Visual Place Recognition},
	author={Shunpeng Chen and Changwei Wang and Rongtao Xu and Peixingtian and yukun Song and Jinzhou Lin and Wenhao Xu and jingyizhang and Li Guo and Shibiao Xu},
	booktitle={The 14 International Conference on Learning Representations},
	year={2026}
}

@inproceedings{Mip-splatting,
	title={Mip-splatting: Alias-free 3d gaussian splatting},
	author={Yu, Zehao and Chen, Anpei and Huang, Binbin and Sattler, Torsten and Geiger, Andreas},
	booktitle={Proceedings of the IEEE/CVF conference on computer vision and pattern recognition},
	pages={19447--19456},
	year={2024}
}

@article{igelio,
	title={IGE-LIO: Intensity gradient enhanced tightly coupled LiDAR-inertial odometry},
	author={Chen, Ziyu and Zhu, Hui and Yu, Biao and Jiang, Chunmao and Hua, Chen and Fu, Xuhui and Kuang, Xinkai},
	journal={IEEE Transactions on Instrumentation and Measurement},
	volume={73},
	pages={1--11},
	year={2024},
	publisher={IEEE}
}

@article{mmdslam,
	title={MMD-SLAM: Structure-Enhanced Multi-Meta Gaussian Distribution-Guided Visual SLAM},
	author={Zhu, Fan and Chen, Ziyu and Liu, Peichen and Zhao, Yifan and Xu, Zhisong and Zhu, Hui and Zhou, Hongxing and Liu, Sixun and Jiang, Chunmao},
	journal={arXiv preprint arXiv:2606.19874},
	year={2026}
}

@article{gof,
	title={Gaussian opacity fields: Efficient adaptive surface reconstruction in unbounded scenes},
	author={Yu, Zehao and Sattler, Torsten and Geiger, Andreas},
	journal={ACM Transactions on Graphics (ToG)},
	volume={43},
	number={6},
	pages={1--13},
	year={2024},
	publisher={ACM New York, NY, USA}
}

@article{faiss,
	title={The faiss library},
	author={Douze, Matthijs and Guzhva, Alexandr and Deng, Chengqi and Johnson, Jeff and Szilvasy, Gergely and Mazar{\'e}, Pierre-Emmanuel and Lomeli, Maria and Hosseini, Lucas and J{\'e}gou, Herv{\'e}},
	journal={IEEE Transactions on Big Data},
	year={2025},
	publisher={IEEE}
}

@inproceedings{droid-splat,
	title={DROID-Splat Combining end-to-end SLAM with 3D Gaussian Splatting},
	author={Homeyer, Christian and Begiristain, Leon and Schn{\"o}rr, Christoph},
	booktitle={Proceedings of the IEEE/CVF International Conference on Computer Vision},
	pages={2767--2777},
	year={2025}
}

@article{framevggt,
	title={FrameVGGT: Frame Evidence Rolling Memory for Streaming VGGT},
	author={Xu, Zhisong and Oishi, Takeshi},
	journal={arXiv preprint arXiv:2603.07690},
	year={2026}
}

@inproceedings{Eigenplaces,
	title={Eigenplaces: Training viewpoint robust models for visual place recognition},
	author={Berton, Gabriele and Trivigno, Gabriele and Caputo, Barbara and Masone, Carlo},
	booktitle={Proceedings of the IEEE/CVF International Conference on Computer Vision},
	pages={11080--11090},
	year={2023}
}

@article{radegs,
	title={Rade-gs: Rasterizing depth in gaussian splatting},
	author={Zhang, Baowen and Fang, Chuan and Shrestha, Rakesh and Liang, Yixun and Long, Xiao-Xiao and Tan, Ping},
	journal={ACM Transactions on Graphics},
	volume={45},
	number={2},
	pages={1--14},
	year={2026},
	publisher={ACM New York, NY}
}

@InProceedings{splatslam,
	author    = {Sandstr\"om, Erik and Zhang, Ganlin and Tateno, Keisuke and Oechsle, Michael and Niemeyer, Michael and Zhang, Youmin and Patel, Manthan and Van Gool, Luc and Oswald, Martin and Tombari, Federico},
	title     = {Splat-SLAM: Globally Optimized RGB-only SLAM with 3D Gaussians},
	booktitle = {Proceedings of the IEEE/CVF Conference on Computer Vision and Pattern Recognition (CVPR) Workshops},
	month     = {June},
	year      = {2025},
	pages     = {1686-1697}
}

\end{document}